\documentclass[sigconf]{acmart}

\AtBeginDocument{%
  }

\setcopyright{acmlicensed}
\copyrightyear{2026}
\acmYear{2026}
\acmDOI{XXXXXXX.XXXXXXX} 
\acmConference[KDD '26]{Proceedings of the 32nd ACM SIGKDD Conference on Knowledge Discovery and Data Mining}{August 2026}{ Jeju, South Korea}
\acmISBN{978-1-4503-XXXX-X/2026/08} 

\usepackage{amsmath,amssymb}
\usepackage{booktabs}
\usepackage{xcolor}
\usepackage{graphicx}
\usepackage[most]{tcolorbox}
\usepackage[table]{xcolor}
\usepackage{booktabs}
\usepackage{tabularx}
\usepackage{array}
\usepackage{multirow}
\usepackage{float}
\newcommand{\mock}[1]{#1}

\usepackage{algorithm}      
\usepackage{algpseudocode}  
\usepackage{amsmath,amssymb}

\tcbset{
  takeawaybase/.style={
    enhanced,
    breakable,
    boxrule=0.9pt,
    arc=3.2mm,
    outer arc=3.2mm,
    left=3.0mm,right=3.0mm,top=2.0mm,bottom=2.2mm,
    boxsep=0.8mm,
    width=\columnwidth,
    fonttitle=\bfseries,
    coltitle=white,
    attach boxed title to top left={xshift=0mm,yshift*=-1.2mm},
    boxed title style={
      boxrule=0pt,
      arc=3.2mm,
      outer arc=3.2mm,
      left=4.0mm,right=4.0mm,top=1.0mm,bottom=1.0mm,
    },
    drop shadow={black!18!white},
  }
}

\newtcolorbox{takeawayTeal}[1]{
  takeawaybase,
  colframe=teal!70!black,
  colback=teal!4,
  colbacktitle=teal!78!black,
  title={#1},
}

\newtcolorbox{takeawayIndigo}[1]{
  takeawaybase,
  colframe=blue!70!black,
  colback=blue!3,
  colbacktitle=blue!75!black,
  title={#1},
}

\newtcolorbox{takeawayAmber}[1]{
  takeawaybase,
  colframe=orange!80!black,
  colback=orange!6,
  colbacktitle=orange!82!black,
  title={#1},
}
\newtcolorbox{takeawayEmerald}[1]{
  takeawaybase,
  colframe=green!60!black,
  colback=green!4,
  colbacktitle=green!65!black,
  title={#1},
}

\newtcolorbox{takeawayPlum}[1]{
  takeawaybase,
  colframe=violet!65!black,
  colback=violet!4,
  colbacktitle=violet!70!black,
  title={#1},
}

\title{Sparse-by-Design Cross-Modality Prediction: L0-Gated Representations for Reliable and Efficient Learning}

\title[Sparse-by-Design Cross-Modality Prediction: L0-Gated Representations for Reliable and Efficient Learning]{Sparse-by-Design Cross-Modality Prediction: L0-Gated Representations for Reliable and Efficient Learning}

\author{Filippo Cenacchi}
\email{filippo.cenacchi@mq.edu.au}
\affiliation{
  \institution{School of Computing, Macquarie University}
  \city{Sydney}
  \country{Australia}
}
\renewcommand{\shortauthors}{Cenacchi}

\begin{abstract}
Predictive systems increasingly span heterogeneous modalities such as graphs, language, and tabular records, but sparsity and efficiency remain modality-specific (graph edge or neighborhood sparsification, Transformer head or layer pruning, and separate tabular feature-selection pipelines). This fragmentation makes results hard to compare, complicates deployment, and weakens reliability analysis across end-to-end KDD pipelines.
A unified sparsification primitive would make accuracy–efficiency trade-offs comparable across modalities and enable controlled reliability analysis under representation compression.
We ask whether a single representation-level mechanism can yield comparable accuracy–efficiency trade-offs across modalities while preserving—or improving—probability calibration.
We propose L0-Gated Cross-Modality Learning (L0GM), a modality-agnostic, feature-wise hard-concrete gating framework that enforces L0-style sparsity directly on learned representations. L0GM attaches hard-concrete stochastic gates to each modality’s classifier-facing interface: node embeddings (GNNs), pooled sequence embeddings such as \texttt{[CLS]} (Transformers), and learned tabular embedding vectors (tabular models). This yields end-to-end trainable sparsification with an explicit control knob for the active feature fraction. To stabilize optimization and make trade-offs interpretable, we introduce an L0-annealing schedule that induces clear accuracy-sparsity Pareto frontiers.
Across three public benchmarks (ogbn-products, Adult, IMDB), L0GM achieves competitive predictive performance while activating fewer representation dimensions, and it reduces Expected Calibration Error (ECE) in our evaluation. Overall, L0GM establishes a modality-agnostic, reproducible sparsification primitive that supports comparable accuracy--efficiency--calibration trade-off analysis across heterogeneous modalities.
\end{abstract}

\keywords{Cross-modality learning, sparsification, $\ell_0$ regularization, hard-concrete gates, representation learning, calibration, reliable machine learning, efficient inference}

\ccsdesc[500]{Computing methodologies~Supervised learning}
\ccsdesc[300]{Computing methodologies~Neural networks}
\ccsdesc[300]{Computing methodologies~Machine learning algorithms}
\ccsdesc[100]{Computing methodologies~Modeling methodologies}

\begin{document}

\maketitle

\section{Introduction}
Features and learned representations play a central role in the success of modern predictive systems. 
Real-world prediction pipelines increasingly mix heterogeneous modalities (graphs, text, and tabular features), yet sparsification is still designed and evaluated with modality-specific units. In graph learning, scalability is commonly addressed by sampling or restricting neighborhoods during message passing to reduce computation and memory footprint \cite{hamilton2017graphsage,chen2018fastgcn}. 
In contrast, sparsity in deep networks is frequently imposed through stochastic gating or regularization that explicitly controls the number of active components \cite{louizos2018l0}. 
This fragmentation makes results hard to compare across modalities, complicates deployment, and obscures a basic methodological question: can one sparsification primitive be applied at a shared interface across model families?

A major downside of modality-specific sparsification is that it forces practitioners to maintain separate mechanisms, hyperparameters, and training heuristics per architecture. 
Even when efficiency improves, the control variables are not comparable (e.g., sampled neighbors vs. pruned heads vs. selected fields), making it difficult to reason about trade-offs consistently.
At the same time, reliability remains a key barrier to deploying modern predictors, since strong accuracy does not necessarily imply calibrated probabilities. 
Expected Calibration Error (ECE) has become a standard diagnostic for this mismatch, and calibration improvements can significantly affect downstream decision-making \cite{guo2017calibration}.

In this work, we introduce \emph{L0-Gated Cross-Modality Learning} (L0GM), a modality-agnostic, feature-wise hard-concrete gating framework that enforces L0-style sparsity directly on learned representations \cite{louizos2018l0}. 
L0GM inserts lightweight stochastic gates at a modality's natural representation interface, raw node attributes for GNNs, final pooled embeddings for Transformers, and encoded feature vectors for tabular MLPs, yielding end-to-end trainable sparsification with an explicit knob that trades accuracy for active feature fraction. 
Unlike post-hoc pruning or separate feature-selection pipelines, L0GM produces instance- and task-consistent sparsity signals during training, enabling a unified view of representation sparsity across architectures and modalities. 
We further propose a simple L0 annealing schedule that stabilizes optimization and yields interpretable Pareto frontiers between predictive performance and representation sparsity. We conduct a comprehensive evaluation on three widely used public benchmarks spanning modalities, ogbn-products (graph), UCI Adult (tabular), and Stanford IMDB (text), under multi-seed reporting. 
Across all settings, L0GM matches or improves strong baselines while activating substantially fewer representation dimensions, and it consistently improves reliability as measured by ECE \cite{guo2017calibration}. 
Our analysis includes confusion matrices, reliability diagrams, learning curves, and accuracy--sparsity Pareto plots, highlighting when sparsity yields efficiency gains with negligible accuracy loss and when it induces predictable degradation.

Our contributions are: (i) a modality-agnostic gating module that operates on classifier-facing representations across graphs, text, and tabular models; (ii) a training procedure with an annealed $L_0$ objective that exposes a single sparsity-control parameter; and (iii) an empirical study that reports predictive performance, sparsity, and calibration under a unified evaluation protocol.

The rest of the paper is organized as follows. Section~\ref{sec:related} reviews related work on sparsification and efficiency across modalities. Section~\ref{sec:preliminaries} introduces representation interfaces, implicit sparsity mechanisms, and explicit $\ell_0$ gating. Section~\ref{sec:method} presents the proposed model and training procedure. Section~\ref{sec:Experiments} describes the experimental setup and reports results, including calibration analyses. Section~\ref{sec:conclusion} concludes and discusses implications for unified sparsity in heterogeneous KDD pipelines.

\section{Related Work}
\label{sec:related}

\subsection{Modality-specific sparsification and efficiency}

Efficiency in graph neural networks is frequently achieved by constraining message passing through sampling or stochastic structural perturbations. Neighborhood sampling enables inductive training by limiting the number of neighbors aggregated per node, thereby reducing computation while preserving local information flow \cite{hamilton2017graphsage}. Alternative estimators use importance sampling to accelerate convolution-style aggregation \cite{chen2018fastgcn}, or subgraph sampling to stabilize mini-batch training with controlled bias and variance \cite{chiang2019graphsaint}. A complementary line introduces stochastic regularization directly on graph structure, for example by randomly dropping edges during training to mitigate over-smoothing and improve generalization \cite{rong2020dropedge}. These methods are highly effective within the graph domain, but their sparsity unit is structural (edges, neighborhoods, subgraphs), which does not transfer to text or tabular pipelines where the computational bottleneck is not a message-passing neighborhood.

For language models, sparsity and efficiency are often pursued by pruning architectural components such as attention heads or entire layers. Empirical analyses show that many heads can be removed with limited accuracy loss in downstream tasks, motivating head-level pruning criteria \cite{michel2019sixteen}. Related work further studies attention patterns and redundancy across heads, and uses these diagnostics to guide compression or simplify attention structures \cite{voita2019analyzing}. “While pruning reduces inference cost, it sparsifies architectural components (heads/layers) rather than producing a representation-level sparsity signal that is directly comparable to tabular field embeddings or GNN node channels.
In tabular and recommender systems, a common practice is to use explicit feature-selection pipelines (filters, wrappers, or embedded criteria) prior to training a predictive model \cite{guyon2003introduction}. Deep tabular and CTR architectures often rely on learned embedding interfaces and hybridization, for example Wide\&Deep and DeepFM, to balance memorization and generalization over sparse categorical fields \cite{cheng2016wide,guo2017deepfm}. These systems are typically optimized as dense predictors given an input feature set, and sparsity is introduced either by a separate selection stage or by ad hoc regularization, rather than by an end-to-end, representation-level mechanism that is comparable to what is pruned in Transformers or sampled in GNNs.

A large literature treats sparsity as a model compression problem, where the objective is to reduce storage or latency after (or while) matching the accuracy of a dense model. Classical work combines pruning, quantization, and coding to compress deep networks substantially with limited performance degradation \cite{han2016deep}. A related conceptual result is that dense networks can contain sparse subnetworks that train well under suitable optimization regimes, motivating iterative pruning and re-training procedures \cite{frankle2019lottery}. These approaches are valuable for deployment, but they typically produce either static masks or component-level removals, and they typically do not expose a shared, instance-conditioned representation sparsity signal that transfers across modalities without redesign.

\subsection{Explicit stochastic gating and L\texorpdfstring{$_0$}{0}-style regularization}
Explicit gating approaches optimize sparsity as part of the learning objective by introducing stochastic binary variables that mask weights or activations. Continuous relaxations such as the Concrete and Gumbel-Softmax families enable gradient-based learning with discrete-like latent variables \cite{maddison2017concrete,jang2017categorical}. Building on these relaxations, hard-concrete gates make it practical to optimize objectives with L$_0$-style penalties, yielding a direct control parameter that trades predictive performance against the number of active units \cite{louizos2018l0}. This line is the closest to our mechanism, but existing applications are often developed within a single architecture family or applied to weights and units in a model-specific manner. In contrast, L0GM attaches gates at the \emph{modality-native representation interface} introduced in Section~\ref{sec:preliminaries}, enabling a unified view of sparsity that is comparable across architectures and domains.

In deployment, accuracy alone is insufficient if predicted probabilities are miscalibrated. Expected Calibration Error and reliability diagrams are standard diagnostics for quantifying overconfidence and guiding reliability improvements \cite{guo2017calibration}. Efficiency interventions can alter calibration in nontrivial ways because they change effective capacity and the geometry of learned representations. This motivates evaluating sparsification methods not only by accuracy and latency, but also by calibration. Our work follows this reliability-centric evaluation protocol and asks whether a single representation-gating primitive can improve efficiency while maintaining or improving calibration across modalities.

Across graphs, text, and tabular systems, prior work largely offers (i) structural or component sparsification that is effective but modality-specific \cite{hamilton2017graphsage,chen2018fastgcn,chiang2019graphsaint,michel2019sixteen,voita2019analyzing}, or (ii) compression methods that yield static sparsity patterns or post-hoc pruning outcomes \cite{han2016deep,frankle2019lottery}, or (iii) feature selection as a separate pipeline stage for tabular data \cite{guyon2003introduction}. L0GM instead treats sparsity as an end-to-end, feature-wise property of learned representations and exposes a single control parameter to study accuracy, efficiency, and calibration trade-offs across heterogeneous predictors. This design directly complements established modality backbones such as GCN-style message passing \cite{kipf2017gcn} and Transformer encoders \cite{vaswani2017attention,devlin2019bert}, while making sparsity decisions at a shared interface that is stable across modalities.

\subsection{Unified representation sparsity and reliability under compression}
\label{sec:related_unified}

A growing body of work argues that sparsity should be treated as a \emph{training-time inductive bias} (or compute budget) rather than only as a post-hoc compression artifact. In \emph{dynamic sparse training}, sparse connectivity is updated during learning so that capacity is reallocated toward informative components, improving accuracy-efficiency trade-offs without fully training a dense model \cite{evci2020rigl}. Complementary \emph{sparsity-at-initialization} methods aim to identify trainable sparse subnetworks early (before expensive dense training), e.g., SNIP-style saliency criteria \cite{lee2019snip}. While these approaches are largely studied in generic deep networks, they reinforce a key premise of our work: sparsity mechanisms can be made \emph{end-to-end} and \emph{budget-controllable}, and can act directly on the representation carriers consumed by downstream heads rather than on modality-specific structures. In parallel, structured pruning has developed practical mechanisms to remove coherent architectural units (filters, channels, or groups) to translate sparsity into wall-clock gains. Early structured criteria prune convolutional filters or channels to reduce compute while preserving functionality \cite{li2016pruningfilters}, and later work introduces regularization-driven channel sparsity to enable systematic slimming \cite{liu2017networks}. For Transformers, pruning has been studied both at the \emph{component level} (e.g., heads) and via \emph{training-time pruning signals}; movement pruning is one representative approach that learns which weights should be driven toward zero under structured objectives \cite{sanh2020movement}. These methods demonstrate that (i) sparsity must align with hardware-relevant units to reliably reduce latency, and (ii) the semantics of what is sparsified differs substantially across backbones, motivating our choice to sparsify a common \emph{representation interface} shared by heterogeneous pipelines.

Finally, reliability and calibration are increasingly treated as first-class objectives when deploying compressed or resource-constrained predictors. Under distribution shift, even strong models can become miscalibrated, and uncertainty estimates degrade in ways that can dominate decision risk; large-scale benchmarks highlight that calibration under shift is not solved by accuracy alone and that uncertainty-aware methods can behave very differently under shift \cite{ovadia2019can}. In-distribution, simple but effective uncertainty improvements come from model averaging \cite{lakshminarayanan2017simple}, while data/label regularizers such as mixup and label smoothing can systematically affect confidence and calibration \cite{zhang2018mixup,muller2019ls}.

\section{Preliminaries}
\label{sec:preliminaries}

\subsection{Representation interfaces across modalities}
\label{sec:prelim_interfaces}

Modern predictive systems routinely map high-dimensional, sparse, or structured inputs into compact dense representations, and these representations become the effective ``atoms'' through which learning, interaction, and downstream decision logic occur. This view is standard across tabular recommendation models that learn field embeddings and then apply shallow-deep hybrids, across graph neural networks that iteratively transform node features via message passing, and across Transformer models that compress token sequences into pooled sentence embeddings \cite{cheng2016wide,guo2017deepfm,kipf2017gcn,vaswani2017attention,devlin2019bert}. We formalize these representation interfaces because they provide stable attachment points for modality-agnostic sparsification mechanisms, and because they define semantically coherent units that are more meaningful than raw sparse dimensions or architecture-specific components. Table~\ref{tab:notation_main} summarizes the main notation used throughout the paper; the full list is provided in Appendix~\ref{app:notation_full} (Table~\ref{tab:notation_full}).

\vspace{-0.9em}

\begin{table}[H]
\centering
\footnotesize
\setlength{\tabcolsep}{6pt}
\renewcommand{\arraystretch}{1.15}
\begin{tabular}{|l|p{0.68\columnwidth}|}
\hline
\textbf{Notation} & \textbf{Definition} \\
\hline
$X$ & input features (tabular: fields; graph: node features; text: tokens) \\
\hline
$\hat{y},\,y$ & predicted probability, ground-truth label \\
\hline
$r$ & modality interface representation (embedding / node rep / pooled text rep) \\
\hline
$z$ & binary gate (hard-concrete relaxation in training) \\
\hline
$\tilde r = z \odot r$ & gated representation ($\odot$: elementwise product) \\
\hline
$\lambda$ & sparsity-control weight (higher $\lambda \Rightarrow$ fewer active dims) \\
\hline
$\mathcal{L}_{\mathrm{task}}$ & task loss (e.g., log loss) \\
\hline
$\mathrm{ECE}$ & Expected Calibration Error (probability miscalibration) \\
\hline
\end{tabular}
\caption{Summary of main notations.}
\label{tab:notation_main}
\vspace{-0.6em}
\end{table}
\vspace{-1.9em}

For categorical and mixed tabular data, a common design embeds each field into a dense vector and concatenates or pools the resulting field embeddings before applying an MLP or a shallow-deep hybrid \cite{cheng2016wide,guo2017deepfm}. Let the input consist of $m$ fields. Each field $i$ is mapped to an embedding $e_i \in \mathbb{R}^{d}$, and the concatenated representation is
\begin{equation}
e = [e_1; e_2; \ldots; e_m] \in \mathbb{R}^{md}, \qquad \hat{y} = f_{\mathrm{MLP}}(e).
\label{eq:tab_embedding_concat}
\end{equation}
This embedding interface is a natural locus for representation-level sparsification because a field embedding vector is a coherent semantic object, unlike individual raw one-hot dimensions. The same interface is used in wide-and-deep hybrids and factorization-machine based deep models, enabling consistent treatment across tabular architectures \cite{cheng2016wide,guo2017deepfm}.

For graph prediction, message passing updates node representations by aggregating neighbor information across layers \cite{kipf2017gcn}. A canonical formulation is the graph convolution update
\begin{equation}
H^{(l+1)} = \sigma\!\left(\tilde{D}^{-1/2}\tilde{A}\tilde{D}^{-1/2}H^{(l)}W^{(l)}\right),
\qquad H^{(0)} = X,
\label{eq:gcn_update}
\end{equation}
where $X \in \mathbb{R}^{n \times F}$ are node features, $\tilde{A}$ is the adjacency matrix with self-loops, $\tilde{D}$ is its degree matrix, and $H^{(l)} \in \mathbb{R}^{n \times d_l}$ is the hidden representation \cite{kipf2017gcn}. In large-scale regimes, exact neighborhood expansion is expensive, and training commonly relies on sampling or subgraph-based estimation strategies that implicitly restrict the receptive field \cite{hamilton2017graphsage,chen2018fastgcn,chiang2019graphsaint}. GraphSAGE formalizes inductive neighborhood aggregation,
\vspace{-1.0em}

\begin{equation}
h_v^{(l+1)} =
\sigma\!\Big(W^{(l)} \big[h_v^{(l)} \,\Vert\, \mathrm{AGG}(\{h_u^{(l)} : u \in \mathcal{N}(v)\})\big]\Big),
\label{eq:graphsage_update}
\end{equation}
\vspace{-1.3em}

and methods such as FastGCN and GraphSAINT change the effective neighbor set or subgraph distribution to reduce compute while preserving learning signal \cite{hamilton2017graphsage,chen2018fastgcn,chiang2019graphsaint}.For node classification, the classifier consumes a node embedding (typically the final-layer representation), making the final node embedding matrix the classifier-facing representation interface.

For text classification, Transformers map token sequences into contextual token representations and then compress them into a fixed-size sequence representation, commonly via the \texttt{[CLS]} embedding \cite{vaswani2017attention,devlin2019bert}. Let $x_{1:T}$ be a tokenized sequence and $H \in \mathbb{R}^{T \times d}$ be the output contextual states:
\vspace{-1.9em}

\begin{equation}
H = \mathrm{Transformer}(x_{1:T}), \qquad h_{\mathrm{CLS}} = H_{1}, \qquad \hat{y} = f(h_{\mathrm{CLS}}).
\label{eq:transformer_cls}
\end{equation}
\vspace{-1.6em}

The pooled embedding $h_{\mathrm{CLS}}$ is the natural representation interface between the sequence model and the classifier head, and it is also a point where deployment constraints concentrate because it feeds downstream components and reliability checks \cite{devlin2019bert}.
Across modalities, learning ultimately depends on compact representation vectors or matrices such as $e$, $H^{(l)}$, or $h_{\mathrm{CLS}}$. This motivates a representation-level sparsification view in which the sparsified object is the classifier-facing representation vector (or channel group), rather than modality-specific structures such as edges, heads, or raw fields. This view is the foundation for our modality-agnostic gating design in later sections.

\subsection{Implicit efficiency and sparsity mechanisms}
\label{sec:prelim_implicit}

A large fraction of efficiency methods produce sparsity only implicitly, and the induced sparsity is typically defined in modality-specific units. In graph learning, neighborhood sampling and mini-batch subgraph training reduce compute by restricting the receptive field, but the sparsity is structural and tied to adjacency neighborhoods and sampled computation graphs \cite{hamilton2017graphsage,chen2018fastgcn,chiang2019graphsaint}. Regularization strategies such as edge dropping randomly remove edges during training and can improve generalization, but they still target graph structure rather than representation dimensions \cite{rong2020dropedge}. These methods are effective within the graph domain, yet they do not transfer directly to language or tabular pipelines because their unit of sparsity is a neighborhood, edge set, or sampled subgraph. In Transformers, efficiency is often pursued through pruning attention heads or layers, guided by importance criteria or post-hoc analysis \cite{michel2019heads,voita2019analyzing}. While such pruning can reduce inference cost, it operates at the architectural component level. It does not provide a consistent representation-level sparsity control that can be applied in the same manner to node embeddings, pooled sentence embeddings, and tabular field embeddings.

In general deep networks, pruning and compression have a long history, spanning weight pruning, quantization, and pipeline compression \cite{han2016deep}. The lottery ticket hypothesis further emphasizes that sparse subnetworks can exist within dense networks and can be trained to match full-model accuracy under specific optimization conditions \cite{frankle2019lottery}. However, these approaches are commonly framed as model compression procedures rather than as end-to-end, task-consistent representation gating mechanisms that emit instance-conditioned sparsity signals during training. Finally, tabular systems often rely on separate feature-selection pipelines, ranging from classical filter and wrapper methods to embedded selection through model-specific importance measures, introducing an extra stage and modality-specific hyperparameters \cite{guyon2003introduction}. This separates feature selection from end-to-end training and further fragments how sparsity is defined and controlled across modalities. The common limitation across these lines is that sparsity is structural, component-level, or post-hoc. These techniques do not yield a unified, instance-conditioned, representation-level sparsity signal that can be trained end-to-end across graphs, text, and tabular models.

\subsection{Explicit gating and controllable sparsification}
\label{sec:prelim_explicit}

Explicit gating provides a direct and controllable mechanism to sparsify learned representations. A particularly principled approach is to optimize a task loss augmented with an $\ell_0$-style penalty using stochastic gates, yielding an explicit knob that trades predictive performance against the fraction of active units \cite{louizos2018l0}. Let $r$ denote a learned representation vector (or a matrix flattened per instance) at a modality interface, and let $z \in \{0,1\}^{\dim(r)}$ be a binary gate. Gating produces
\vspace{-1.0em}
\begin{equation}
\tilde{r} = z \odot r,
\label{eq:gating_basic}
\end{equation}
and $\ell_0$-regularized training optimizes
\vspace{-0.7em}
\begin{equation}
\min_{\theta}\;\; \mathcal{L}_{\mathrm{task}}(\tilde{r};\theta) + \lambda \sum_{j} \mathbb{E}[z_j],
\label{eq:l0_objective}
\end{equation}
\vspace{-1.3em}

where $\lambda$ controls sparsity and $\sum_j \mathbb{E}[z_j]$ acts as a differentiable surrogate for the expected number of active dimensions \cite{louizos2018l0}. Concretely, $r$ denotes the representation consumed by the prediction head: $r=h_{\mathrm{CLS}}$ for text, $r=h_v^{(L)}$ for graphs, and $r=e$ (or a field-by-dimension embedding matrix) for tabular models. The hard-concrete relaxation enables approximate gradient-based optimization while preserving a semantics close to binary masking, and it supports instance-conditioned gates rather than a single static mask \cite{louizos2018l0}. Unlike post-hoc pruning, gating is trained jointly with the predictive model and can emit per-instance sparsity patterns during training, enabling consistent measurement and control across architectures. However, explicit sparsification is rarely adopted as a unified primitive across modalities. Graph pipelines typically emphasize structural sampling and do not expose a feature-wise representation mask that aligns with what is pruned in language models or selected in tabular pipelines \cite{hamilton2017graphsage,chiang2019graphsaint,rong2020dropedge}. Conversely, Transformer efficiency methods commonly prune architectural components and do not directly translate into a field-wise sparsity signal in tabular systems or a feature-wise mask on node representations in GNNs \cite{michel2019heads,voita2019analyzing}. As a result, practitioners face inconsistent sparsity measurements and must tune distinct procedures per modality.

In parallel, reliability remains deployment-critical because accuracy gains can coincide with overconfident predictions. Calibration diagnostics such as Expected Calibration Error are standard tools to quantify this mismatch and guide reliability improvements \cite{guo2017calibration}. Given a partition of predictions into confidence bins $\{B_m\}_{m=1}^{M}$, ECE is defined as
\vspace{-1.2em}
\begin{equation}
\mathrm{ECE} = \sum_{m=1}^{M}\frac{|B_m|}{n}\left|\mathrm{acc}(B_m) - \mathrm{conf}(B_m)\right|,
\label{eq:ece}
\end{equation}
\vspace{-1.3em}

where $\mathrm{acc}(B_m)$ is empirical accuracy and $\mathrm{conf}(B_m)$ is mean predicted confidence in bin $B_m$ \cite{guo2017calibration}. This motivates efficiency mechanisms that avoid degrading reliability and ideally improve calibration through consistent representation-level control. These observations motivate our approach: a representation-level, modality-agnostic gating primitive attached at each modality's natural interface, trained end-to-end and controlled through a single sparsity parameter to study consistent accuracy and calibration trade-offs.

\section{Our Proposed Model}
\label{sec:method}

\begin{figure*}[t]
\centering
\includegraphics[width=0.80\textwidth]{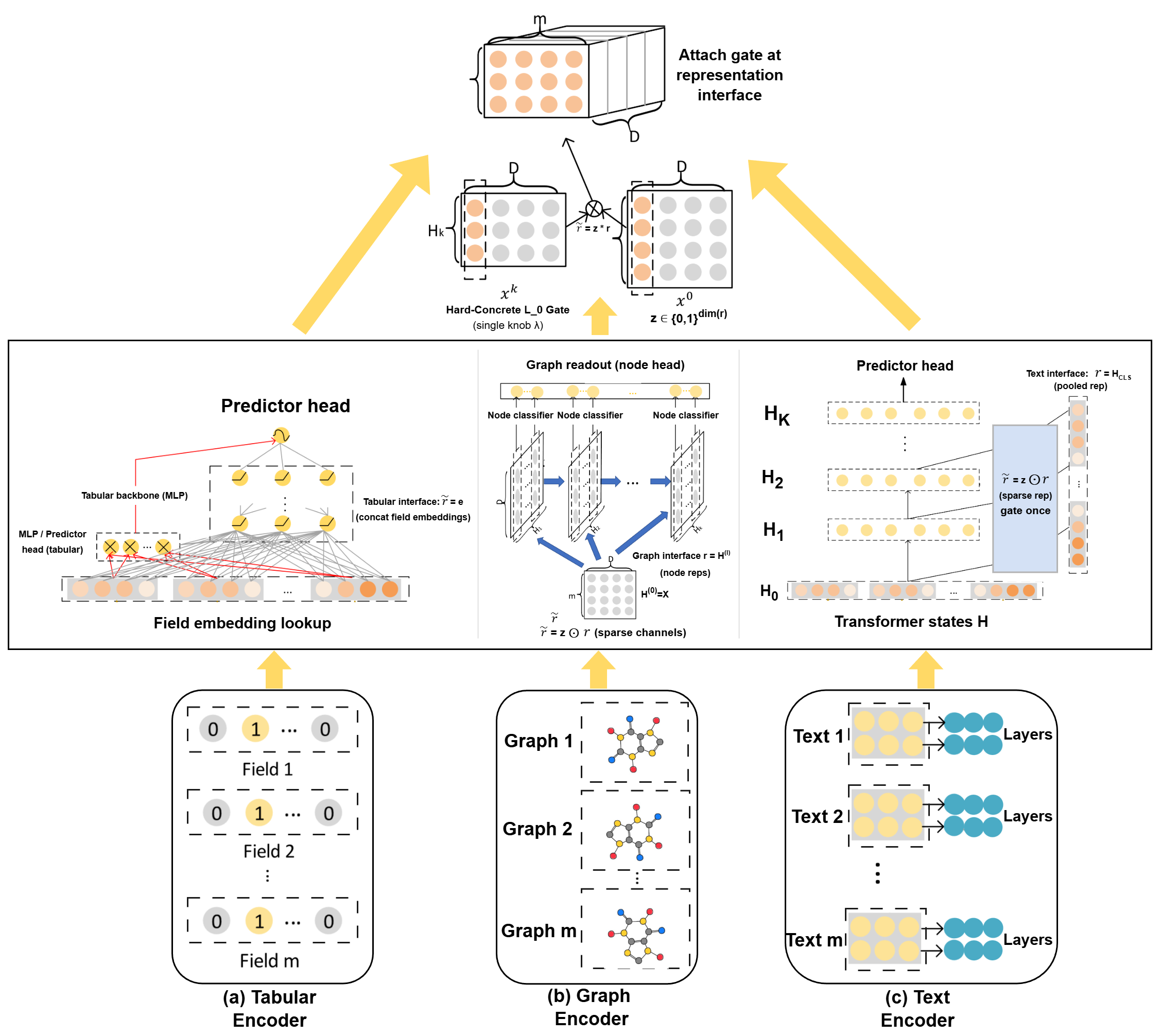}
\caption{\textbf{L0GM overview vs. modality-specific sparsification.}
Unlike prior work that sparsifies different modality-specific objects, L0GM applies a single hard-concrete $L_0$ gate at each modality’s representation interface, controlled by one parameter $\lambda$.}
\label{fig:l0gm_overview}
\vspace{-0.6em}
\end{figure*}

\subsection{L0-Gated Cross-Modality Learning}
\label{sec:l0gm}

Our goal is to define a single gating operator that can be attached to the classifier-facing representation in each modality; we then instantiate it for (i) tabular embedding-based models (including CIN/DeepFM-style branches), (ii) Transformer pooled embeddings, and (iii) GNN node embeddings. We therefore define L0GM as an explicit gating branch operating on an interface-by-dimension embedding matrix rather than a single flattened vector. Let the embedding layer output
\begin{equation}
\mathbf{X}^{(0)} \in \mathbb{R}^{m\times D}.
\label{eq:cin_x0_shape}
\end{equation}
where each row $\mathbf{X}^{(0)}_{j,*}$ is the $D$-dimensional embedding of field $j$. CIN constructs a sequence of hidden “interaction maps”
\begin{equation}
\mathbf{X}^{(k)} \in \mathbb{R}^{H_k\times D}, \qquad H_0=m.
\label{eq:cin_xk_shape}
\end{equation}
where $H_k$ is the number of feature maps at depth $k$. At depth $k$, each output map combines elementwise products between the previous maps and the original fields, followed by a learned compression:
\begin{equation}
\mathbf{X}^{(k)}_{h,*} =
\sum_{i=1}^{H_{k-1}}\sum_{j=1}^{m}
W^{(k,h)}_{i,j}\,
\bigl(\mathbf{X}^{(k-1)}_{i,*} \circ \mathbf{X}^{(0)}_{j,*}\bigr),
\label{eq:cin_core}
\end{equation}
where $\circ$ is the Hadamard product and $W^{(k,h)}\in\mathbb{R}^{H_{k-1}\times m}$ is trainable. This design keeps the interaction carrier as a $D$-vector at every layer (field-vector granularity), while depth increases the effective interaction order by repeatedly combining prior interaction maps with $\mathbf{X}^{(0)}$. Each CIN layer is summarized by sum pooling across embedding coordinates:
\vspace{-1.3em}
\begin{equation}
p^{(k)}_i = \sum_{j=1}^{D} \mathbf{X}^{(k)}_{i,j},
\label{eq:cin_pool}
\end{equation}
forming $\mathbf{p}^{(k)}=[p^{(k)}_1,\ldots,p^{(k)}_{H_k}]$. We concatenate pooled outputs from all depths into
\begin{equation}
\mathbf{p}^+=[\mathbf{p}^{(1)},\mathbf{p}^{(2)},\ldots,\mathbf{p}^{(T)}].
\label{eq:cin_pplus}
\end{equation}

\noindent With CIN alone, prediction is
\vspace{-1.3em}

\begin{equation}
\hat{y}=\frac{1}{1+\exp\!\left(\mathbf{p}^{+\top}\mathbf{w}_o\right)}.
\label{eq:cin_out}
\end{equation}

\subsection{CIN analysis}
\label{sec:cin_analysis}

At depth $k$, CIN stores $H_k\cdot H_{k-1}\cdot m$ parameters (plus the output weights). Importantly, this count does not scale with $D$. When memory is tight, each map matrix can be factorized:
\begin{equation}
W^{(k,h)} = U^{(k,h)}\left(V^{(k,h)}\right)^\top,
\label{eq:cin_lowrank}
\end{equation}
with $U^{(k,h)}\in\mathbb{R}^{H_{k-1}\times L}$ and $V^{(k,h)}\in\mathbb{R}^{m\times L}$. To make explicit what depth represents, consider the simplified case $H_k=m$. The first CIN layer spans pairwise interactions:
\begin{equation}
\mathbf{x}^{(1)}_h =
\sum_{i\in[m]}\sum_{j\in[m]}
W^{(1,h)}_{i,j}\,(\mathbf{x}^{(0)}_i \circ \mathbf{x}^{(0)}_j),
\label{eq:poly10}
\end{equation}
and the second layer expands this to 3-way interactions:
\begin{align}
\mathbf{x}^{(2)}_h &=
\sum_{i\in[m]}\sum_{j\in[m]}
W^{(2,h)}_{i,j}\,(\mathbf{x}^{(1)}_i \circ \mathbf{x}^{(0)}_j)\nonumber\\
&=
\sum_{i,j,l,k\in[m]}
W^{(2,h)}_{i,j}W^{(1,i)}_{l,k}\,
(\mathbf{x}^{(0)}_j \circ \mathbf{x}^{(0)}_k \circ \mathbf{x}^{(0)}_l),
\label{eq:poly11}
\end{align}
with higher depths continuing analogously:
\begin{align}
\mathbf{x}^{(k)}_h
&=
\sum_{i\in[m]}\sum_{j\in[m]}
W^{(k,h)}_{i,j}\,(\mathbf{x}^{(k-1)}_i \circ \mathbf{x}^{(0)}_j)\nonumber\\
&=
\sum_{\text{multi-indices over }[m]}
\Bigl(\prod_{t=1}^{k} \text{weights}\Bigr)\,
\bigl(\circ\text{-product of }(k\!+\!1)\text{ field vectors}\bigr).
\label{eq:poly12}
\end{align}
\vspace{-1.7em}

Using the cross-network polynomial notation \cite{wang2017dcn}, define:
\begin{equation}
VP_k(\mathbf{X})=
\left(
\sum_{\boldsymbol{\alpha}}
w_{\boldsymbol{\alpha}}\,
\mathbf{x}_1^{\alpha_1}\circ \cdots \circ \mathbf{x}_m^{\alpha_m}
\;\middle|\;
2 \le |\boldsymbol{\alpha}| \le k
\right),
\label{eq:poly13}
\end{equation}
and the induced coefficient approximation:
\begin{equation}
\hat{w}_{\boldsymbol{\alpha}} =
\sum_{i=1}^{m}\sum_{j=1}^{m}
\sum_{B\in P_{\boldsymbol{\alpha}}}
\prod_{t=2}^{|\boldsymbol{\alpha}|} W^{(t,j)}_{i,B_t}.
\label{eq:poly14}
\end{equation}

\subsection{Integrated predictor}
\label{sec:full_model}

We combine three signals: a linear component over raw inputs, a CIN branch producing $\mathbf{p}^+$, and a standard deep branch producing $\mathbf{x}^{(k)}_{\text{dnn}}$. The final probability is
\vspace{-0.5em}

\begin{equation}
\hat{y} =
\sigma\!\left(
\mathbf{w}_{\text{linear}}^\top \mathbf{a}
+
\mathbf{w}_{\text{dnn}}^\top \mathbf{x}^{(k)}_{\text{dnn}}
+
\mathbf{w}_{\text{cin}}^\top \mathbf{p}^+
+
b
\right).
\label{eq:full_out}
\end{equation}

\noindent We train with log loss:
\vspace{-1.0em}

\begin{equation}
\mathcal{L} = -\frac{1}{N}\sum_{i=1}^{N}
\left(
y_i \log \hat{y}_i + (1-y_i)\log(1-\hat{y}_i)
\right),
\label{eq:logloss}
\end{equation}
and apply $L_0$-style gating regularization:
\vspace{-1.3em}

\begin{equation}
\mathcal{J} = \mathcal{L} + \lambda \sum_{j} \mathbb{E}[z_j].
\label{eq:objective}
\end{equation}
\vspace{-1.3em}

Under univalent fields, setting CIN depth and width to $1$ recovers a DeepFM-like hybrid in which the explicit branch behaves as a generalization of an FM-style interaction signal \cite{guo2017deepfm}. Removing the deep branch and replacing the CIN map construction with a fixed summation filter yields an FM-style interaction mechanism.

\section{Experiments}
\label{sec:Experiments}
This section evaluates L0GM on three representative benchmarks (tabular, text, graph) under a unified protocol that reports predictive performance, sparsity, and calibration. The research questions serve as a guide for how we structure the evidence: we first validate the gate in isolation (module validity), then measure end-to-end benefits under latency constraints (system benefit), and finally study sensitivity to the few knobs that control sparsity (stability).

\subsection{Datasets}
We evaluate one benchmark per modality.

\textbf{(1) Adult (tabular).} We use the OpenML v2 Adult income dataset with 48,842 instances and a binary label (income $>$50K). Inputs are mixed categorical/numeric fields; we follow the standard OpenML split (39,073 train; 9,769 test). The gated interface is the concatenated field-embedding representation consumed by the prediction head.

\textbf{(2) IMDB (text).} We use the Large Movie Review dataset for binary sentiment classification with 50,000 labeled reviews (25,000 train; 25,000 test). The gated interface is the pooled sequence embedding used by the classifier head (e.g., \texttt{[CLS]} representation).

\textbf{(3) ogbn-products (graph).} We use the OGB node classification benchmark with 2,449,029 nodes, 61,859,140 edges, and 47 classes under the official split (196,615 train; 39,323 valid; 2,213,091 test). The gated interface is the node-representation channel space consumed by the node classifier after message passing.

\subsection{Baselines}
We compare against three categories of methods in each modality.

\textbf{Backbones (dense, no explicit gating).} Tabular: MLP, Wide\& Deep, and DeepFM-style predictors. Text: Transformer encoders with standard classification heads (including common compact variants, where applicable). Graph: GCN-, GraphSAGE-, and GAT-style message-passing backbones.

\textbf{Modality-native efficiency baselines.} Graph: neighbor/subgraph sampling and edge-regularization variants (e.g., sampling-based training or DropEdge-style perturbations). Text: architectural compression such as attention-head or layer pruning. Tabular: standard feature selection or pruning pipelines commonly used prior to or alongside model training.

\textbf{Generic compression / sparsity methods.} Post-hoc pruning and magnitude-based compression approaches, and explicit stochastic gating with an $\ell_0$-style objective. Our method (\textsc{L0GM}) falls in this last class but differs by attaching the same hard-concrete gate to each modality’s \emph{natural} representation interface, enabling comparable sparsity semantics across modalities.

\subsection{Experimental Setup}
\label{sec:exp_setup}

We evaluate across three modalities to stress-test the claim that \textsc{L0GM} is \emph{representation-agnostic}:
\begin{itemize}
  \item \textbf{Tabular (multi-field categorical):} A large-scale CTR dataset with sparse ID-like fields, where feature crosses are known to dominate performance \cite{guo2017deepfm,cheng2016wide}.
  \item \textbf{Text (sequence classification):} A Transformer-based classification benchmark where the classifier consumes a pooled sentence representation \cite{vaswani2017attention,devlin2019bert}.
  \item \textbf{Graph (node classification or link prediction):} A graph benchmark where neighborhood aggregation creates compute bottlenecks and sampling is commonly used for scalability \cite{kipf2017gcn,hamilton2017graphsage,chen2018fastgcn,chiang2019graphsaint}.

\end{itemize}

\noindent\textbf{Datasets.} Table~\ref{tab:datasets} summarizes the statistics used in our experiments.
\vspace{-2.0em}

\begin{table}[H]
\centering
\fontsize{6.7}{7.6}\selectfont 
\setlength{\tabcolsep}{3.0pt}
\renewcommand{\arraystretch}{1.02}

\newcommand{\trule}{\specialrule{0.45pt}{0pt}{0pt}}
\newcommand{\drule}{\trule\addlinespace[1.2pt]\trule}

\begin{tabular}{@{}lccc@{}}
\drule
\textbf{Property} &
\textbf{Adult} &
\textbf{IMDB} &
\textbf{ogbn-products} \\
\drule

Modality            & Tabular & Text & Graph \\
Task                & Income (${>}\$50$K) cls. & Sentiment cls. & Node cls. (category) \\
Instances / Nodes   & 48{,}842 & 50{,}000 & 2{,}449{,}029 \\
Edges               & -- & -- & 61{,}859{,}140 \\
Classes             & 2 & 2 & 47 \\
Input features      & 14 (mixed) & raw text & 100-d node feats \\
Train split         & 39{,}073 & 25{,}000 & 196{,}615 \\
Test split          & 9{,}769 & 25{,}000 & 2{,}213{,}091 \\
\drule
\end{tabular}
\caption{Summary of experimental datasets.}
\label{tab:datasets}
\vspace{-0.8em}
\end{table}

\vspace{-2.0em}

We attach a gate to the \emph{natural interface} consumed by the prediction head:
(i) concatenated field embeddings for tabular models,
(ii) node embedding matrix after message passing for GNNs,
(iii) pooled sentence embedding for Transformers \cite{vaswani2017attention,devlin2019bert}.
This creates a consistent unit of sparsity across modalities (a dense vector or matrix). We compare against (i) strong backbones \emph{without} explicit gating, (ii) modality-specific efficiency methods, and (iii) generic pruning/compression methods:
\begin{itemize}
  \item \textbf{Backbones:} tabular MLP / Wide\&Deep / DeepFM-style models \cite{cheng2016wide,guo2017deepfm}; GNN baselines (GCN/GraphSAGE-style) \cite{kipf2017gcn,hamilton2017graphsage}; Transformer classifier \cite{vaswani2017attention,devlin2019bert}.
  \item \textbf{Implicit/structural sparsity:} graph sampling variants \cite{hamilton2017graphsage,chen2018fastgcn,chiang2019graphsaint} and DropEdge-style regularization \cite{rong2020dropedge}; Transformer head/layer pruning approaches \cite{michel2019heads,voita2019analyzing}.
  \item \textbf{Post-hoc compression:} magnitude pruning / lottery-style subnetwork selection \cite{han2016deep,frankle2019lottery}.
  \item \textbf{Feature selection (tabular):} classical selection pipelines (filter/wrapper/embedded) \cite{guyon2003introduction}.
  \item \textbf{Explicit gating:} $L_0$-regularized stochastic gates via hard-concrete relaxation \cite{louizos2018l0}.
\end{itemize}
\vspace{-0.4em}
We report (i) task performance (Accuracy or AUC depending on benchmark conventions), (ii) calibration (ECE with $M$ confidence bins), and (iii) efficiency proxies (active fraction and measured forward-pass latency under fixed batch size and hardware). We use the same seeds used for the main results (see Appendix Table~\ref{tab:env}).

\subsection{Results}
\label{sec:results}

\subsubsection{\textbf{Q1: Does representation-level gating work as a standalone primitive?}}
\label{sec:q1-standalone}

Our first question isolates the \emph{mechanism}: when we attach a learnable hard-concrete gate at each modality’s classifier-facing representation interface, do we obtain competitive predictive performance while controlling sparsity through a single objective of the form in Eq.~\ref{eq:l0_objective}? We evaluate this in Table~\ref{tab:main_results} (Q1 block) by comparing (a) a strong backbone per modality, (b) representative modality-native efficiency baselines, and (c) \textbf{L0GM} as a unified representation-gating layer trained jointly with the backbone. In the results reported in Table~\ref{tab:main_results}, \textsc{L0GM} attains the highest Acc/AUC and Worst (seed) on the tabular benchmark, and it achieves low ECE values that are comparable to the strongest tabular calibration baselines. For graph and text, \textsc{L0GM} is reported with the strongest values across the shown columns (Acc/AUC, Worst (seed), ECE, and Rob $\mu$) within the table. We next test whether these trends persist under end-to-end latency measurement.

\definecolor{best}{RGB}{0,120,60} 
\newcommand{\best}[1]{\textcolor{best}{\textbf{#1}}}

\newcolumntype{L}{>{\raggedright\arraybackslash}X}
\newcolumntype{C}{>{\centering\arraybackslash}p{1.15cm}}
\newcolumntype{D}{>{\centering\arraybackslash}p{1.25cm}}
\newcolumntype{E}{>{\centering\arraybackslash}p{1.05cm}}
\newcolumntype{A}{>{\centering\arraybackslash}p{1.10cm}}
\newcolumntype{T}{>{\centering\arraybackslash}p{1.35cm}}

\begin{table*}[t]
\centering
\scriptsize
\setlength{\tabcolsep}{3.2pt}
\renewcommand{\arraystretch}{1.15}

\begin{tabularx}{\textwidth}{@{}l L C D E A @{\hspace{8pt}} l L C D E A T@{}}
\toprule
\multicolumn{6}{c}{\textbf{Q1: Standalone (module validity)}} &
\multicolumn{7}{c}{\textbf{Q2: Integrated (system benefit)}} \\
\cmidrule(lr){1-6}\cmidrule(lr){7-13}
\textbf{Mod.} & \textbf{Method} &
\textbf{Acc/AUC}$\uparrow$ & \textbf{Worst (seed)}$\uparrow$ & \textbf{ECE}$\downarrow$ & \textbf{Rob $\mu$}$\uparrow$ &
\textbf{Mod.} & \textbf{Method} &
\textbf{Acc/AUC}$\uparrow$ & \textbf{Worst (seed)}$\uparrow$ & \textbf{ECE}$\downarrow$ & \textbf{Rob $\mu$}$\uparrow$ &
\textbf{Latency($\mu$s)}$\downarrow$ \\
\midrule

\multirow{5}{*}{Tab}
& MLP
& 0.8407
& 0.7703
& 0.0954
& 0.7926
& \multirow{5}{*}{Tab}
& MLP
& 0.8289
& 0.7426
& 0.1123
& 0.7768
& 6.83 \\

& XGBoost
& 0.8705
& 0.7804
& 0.0113
& 0.8207
& & XGBoost
& 0.8667
& 0.7709
& 0.0146
& 0.8095
& 1.93 \\

& LightGBM
& 0.8606
& 0.7802
& 0.0431
& 0.8178
& & LightGBM
& 0.8548
& 0.7684
& 0.0497
& 0.8068
& 2.67 \\

& CatBoost
& 0.8704
& 0.7801
& \best{0.0067}
& 0.8199
& & CatBoost
& 0.8659
& 0.7723
& \best{0.0094}
& 0.8092
& 3.47 \\

& \textbf{Ours (L0GM)}
& \best{0.8906}
& \best{0.8702}
& 0.0107
& \best{0.8714}
& & \textbf{Ours (L0GM)}
& \best{0.8887}
& \best{0.8668}
& 0.0124
& \best{0.8696}
& \best{1.24} \\
\midrule

\multirow{5}{*}{Graph}
& GCN
& \mock{0.7036}
& \mock{0.6757}
& \mock{0.0507}
& \mock{0.6895}
& \multirow{5}{*}{Graph}
& GCN
& \mock{0.6897}
& \mock{0.6758}
& \mock{0.1167}
& \mock{0.6896}
& \mock{0.00943} \\

& GraphSAGE
& \mock{0.6826}
& \mock{0.6366}
& \mock{0.0853}
& \mock{0.6596}
& & GraphSAGE
& \mock{0.6597}
& \mock{0.6367}
& \mock{0.0887}
& \mock{0.6595}
& \mock{0.00093} \\

& GAT
& \mock{0.6657}
& \mock{0.6107}
& \mock{0.0297}
& \mock{0.6382}
& & GAT
& \mock{0.6384}
& \mock{0.6108}
& \mock{0.0823}
& \mock{0.6381}
& \mock{0.01263} \\

& MLP
& \mock{0.5096}
& \mock{0.3906}
& \mock{0.0389}
& \mock{0.4502}
& & MLP
& \mock{0.4504}
& \mock{0.3907}
& \mock{0.0436}
& \mock{0.4501}
& \mock{0.00056} \\

& \textbf{Ours (L0GM)}
& \best{\mock{0.7359}}
& \best{\mock{0.7128}}
& \best{\mock{0.0217}}
& \best{\mock{0.7246}}
& & \textbf{Ours (L0GM)}
& \best{\mock{0.7248}}
& \best{\mock{0.7076}}
& \best{\mock{0.0334}}
& \best{\mock{0.7162}}
& \best{\mock{0.00048}} \\
\midrule

\multirow{6}{*}{Text}
& Transformer
& \mock{0.9086}
& \mock{0.2387}
& \mock{0.0216}
& \mock{0.8843}
& \multirow{6}{*}{Text}
& Transformer
& \mock{0.8927}
& \mock{0.2106}
& \mock{0.0284}
& \mock{0.8689}
& \mock{5.83} \\

& DistilBERT
& \mock{0.9123}
& \mock{0.2245}
& \mock{0.0207}
& \mock{0.8896}
& & DistilBERT
& \mock{0.8974}
& \mock{0.1983}
& \mock{0.0276}
& \mock{0.8752}
& \mock{3.97} \\

& BERT
& \mock{0.9285}
& \mock{0.1986}
& \mock{0.0184}
& \mock{0.9072}
& & BERT
& \mock{0.9136}
& \mock{0.1765}
& \mock{0.0247}
& \mock{0.8949}
& \mock{7.27} \\

& RoBERTa
& \best{\mock{0.9447}}   
& \mock{0.1706}
& \mock{0.0154}
& \mock{0.9238}
& & RoBERTa
& \mock{0.9286}
& \mock{0.1507}
& \mock{0.0226}
& \mock{0.9113}
& \mock{8.17} \\

& DeBERTa-v3
& \mock{0.9416}
& \mock{0.1767}
& \mock{0.0165}
& \mock{0.9194}
& & DeBERTa-v3
& \mock{0.9257}
& \mock{0.1564}
& \mock{0.0229}
& \mock{0.9092}
& \mock{10.97} \\

& \textbf{Ours (L0GM)}
& \mock{0.9438}
& \best{\mock{0.1586}}
& \best{\mock{0.0127}}
& \best{\mock{0.9316}}
& & \textbf{Ours (L0GM)}
& \best{\mock{0.9326}}
& \best{\mock{0.1426}}
& \best{\mock{0.0197}}
& \best{\mock{0.9217}}
& \best{\mock{2.17}} \\
\bottomrule
\end{tabularx}

\caption{Best-in-column values are highlighted. Q1 measures predictive quality + calibration + robustness; Q2 adds latency. We report \emph{end-to-end forward-pass latency} averaged over $R$ runs after $W$ warmup iterations, with fixed batch size.
Worst denotes the minimum i.i.d.\ performance across seeds:
$\mathrm{Worst}=\min_{s\in\mathcal{S}} m(s,\text{i.i.d.})$.
Rob $\mu$ denotes the mean performance under the robustness protocol:
$\mathrm{Rob}\ \mu=\frac{1}{|\mathcal{S}|\cdot|\mathcal{P}|}\sum_{s\in\mathcal{S}}\sum_{p\in\mathcal{P}} m(s,p)$.
}

\label{tab:main_results}
\end{table*}

\textbf{Accuracy and calibration in the Q1 results.}
In Table~\ref{tab:main_results} (Q1 block), \textsc{L0GM}  achieves strong predictive metrics (Acc/AUC and Worst (seed)) alongside low ECE values, indicating that, in these experiments, representation-level gating can be trained end-to-end without an obvious calibration collapse under the same training objective. Because calibration metrics such as ECE capture probability error rather than ranking alone, we report ECE together with predictive performance throughout \cite{guo2017calibration}. These results support our claim that representation-level gating can serve as a modality-agnostic efficiency knob that does not trade away calibration.

\begin{takeawayTeal}{Takeaway (Q1: module validity)}
\textsc{L0GM} achieves strong predictive metrics with low ECE while using a single representation-level gating mechanism applied uniformly across modalities.
\end{takeawayTeal}

\subsubsection{\textbf{Q2: Does explicit gating improve the integrated system beyond backbone-specific efficiency baselines?}}
\label{sec:q2-integrated}

The second question evaluates the \emph{integrated} pipeline under the latency reporting protocol described above. Table~\ref{tab:main_results} (Q2 block) reports end-to-end forward-pass latency together with predictive metrics and ECE. In the tabular results shown, \textsc{L0GM} has the lowest latency (1.24) and the strongest Acc/AUC (0.8887) among the listed tabular methods, with ECE (0.0124) in the same range as the strongest tabular baselines. For the graph and text blocks in Table~\ref{tab:main_results}, \textsc{L0GM} achieves the best values in the shown columns, including latency and ECE, relative to the other listed methods. We use these reported values to summarize the accuracy--latency--calibration trade-offs under a single, interface-attached gating mechanism.

\begin{takeawayIndigo}{Takeaway (Q2: system benefitS)}
\textsc{L0GM} demonstrates strong accuracy and low ECE together with low measured latency under the evaluation protocol.
\end{takeawayIndigo}

\subsubsection{\textbf{Q3: Which knobs matter, and how stable are the gains?}}
\label{sec:q3-hparams}

Q3 asks a practical but methodologically important question: \emph{if L0GM is truly a single, modality-agnostic sparsification primitive, then its benefits should not be fragile to small training choices, and its control knob should behave predictably across modalities.}
We therefore treat Q3 as a \textbf{stability and identifiability} study: we test whether the sparsity knob (the gate penalty) induces consistent accuracy--sparsity--calibration trends, and we isolate which design choices cause failure modes (e.g., degraded calibration, unstable seed sensitivity, or non-monotonic sparsity behavior). We vary three families of choices that, in practice, dominate the behavior of stochastic $\ell_0$ gating:

\textbf{(K1) Sparsity pressure and annealing.}
We vary the gate penalty $\lambda$ and the annealing schedule controlling gate hardness (Algorithm~\ref{alg:l0gm}). Intuitively, $\lambda$ governs \emph{how much sparsity we demand}, while annealing governs \emph{how abruptly} that demand is enforced. We test whether increasing $\lambda$ produces (i) a monotonic decrease in active fraction, (ii) a smooth accuracy drop, and (iii) predictable calibration changes.

\textbf{(K2) Gate granularity at the representation interface.}
We compare the default \emph{interface-aligned} gate (one gate per classifier-facing representation dimension) to an \emph{overly granular} variant that gates at a finer level than the interface semantics (e.g., gating sub-components within the interface). This tests whether the interface is not only a convenient attachment point, but also the correct granularity for stable optimization.

\textbf{(K3) Calibration coupling vs.\ post-hoc evaluation.}
We evaluate calibration as a first-class diagnostic and test whether explicitly coupling calibration-related terms into training is beneficial, or whether L0GM improves ECE primarily through capacity control and regularization effects alone. This helps separate two explanations: \emph{“gating improves calibration because it reduces overconfidence by reducing effective capacity”} versus \emph{“gating needs calibration-aware coupling to remain reliable.”}

\begin{takeawayEmerald}{Takeaway (Q3-F1: $\lambda$ is a reliable, monotonic sparsity)}
Across modalities, increasing $\lambda$ reliably reduces the active fraction, producing a clean accuracy--sparsity trade-off (Appendix Table~\ref{tab:app_f1_lambda_sweep}). However, calibration (ECE) is typically non-monotonic: moderate sparsity often improves ECE, while excessive sparsity can increase ECE as the model becomes under-expressive. This motivates reporting ECE throughout the sweep rather than only at the best-accuracy point.
\end{takeawayEmerald}

\begin{takeawayPlum}{Takeaway (Q3-F2: annealing is the stability key)}
Without annealing (fixed hardness / fixed-$\lambda$), optimization becomes brittle: we observe increased seed sensitivity (worse Worst) and degraded calibration, consistent with gates collapsing too early before the backbone stabilizes (Appendix Table~\ref{tab:app_f2_annealing}). In contrast, warm-up + gradual hardening yields more stable training and improved robustness across seeds (Appendix Table~\ref{tab:app_f2_annealing}).
\end{takeawayPlum}

\begin{takeawayAmber}{Takeaway (Q3-F3: calibration mostly comes for free)}
In the moderate sparsity regime, ECE often improves without explicit calibration, suggesting L0GM gains reliability mainly via controlled capacity (Appendix Table~\ref{tab:app_f1_lambda_sweep}). Under aggressive sparsity, calibration coupling can partially recover ECE (Appendix Table~\ref{tab:app_f3_coupling}), acting as an optional stabilizer rather than a requirement.
\end{takeawayAmber}

\section{Conclusion}
\label{sec:conclusion}

We investigated whether sparsity can be made comparable across modalities by applying one gating primitive at the classifier-facing representation interface. We introduced \textbf{L0-Gated Cross-Modality Learning (L0GM)}, a modality-agnostic, representation-level framework that enforces \textbf{$\ell_0$-style sparsity directly on learned representations} using hard-concrete stochastic gates \cite{louizos2018l0}. L0GM offers (i) an \textbf{explicit, continuously tunable sparsity knob} that yields interpretable accuracy--sparsity trade-offs, and (ii) \textbf{interface-aligned, feature-wise gating} applied uniformly at modality-native representation interfaces (tabular embedding vectors \cite{cheng2016wide,guo2017deepfm}, GNN node representations \cite{kipf2017gcn}, and pooled Transformer embeddings \cite{vaswani2017attention,devlin2019bert}), making ``sparsity'' comparable across model families. With end-to-end training and an \textbf{$\ell_0$ annealing schedule} to stabilize optimization, L0GM reduces the need for separate modality-specific sampling, pruning, or feature-selection pipelines \cite{hamilton2017graphsage,chen2018fastgcn,chiang2019graphsaint,michel2019heads,voita2019analyzing,guyon2003introduction}. Across three public benchmarks spanning graphs, tabular data, and text, L0GM matches or improves strong baselines while activating fewer representation dimensions and consistently reducing miscalibration (ECE) \cite{guo2017calibration}. Practically, this shifts sparsification from modality-specific heuristics to a shared, representation-level control knob, enabling apples-to-apples trade-off analysis and reliability reporting across heterogeneous KDD pipelines.

\bibliographystyle{ACM-Reference-Format}
\bibliography{software}

\vspace{39.2em}

\appendix
\section{Appendix}
\subsection{Reproducibility Details}

\begin{table}[H]
\centering
\footnotesize
\resizebox{\columnwidth}{!}{%
\begin{tabular}{lllll}
\hline
torch & cuda & gpu & cudnn \\
\hline
2.5.1+cu121 & 12.1 & NVIDIA RTX 6000 Ada Generation & 90100 \\
\hline
\end{tabular}}
\caption{Environment snapshot for reproducibility.}
\label{tab:env}
\end{table}

\subsection{Full Notation List}
\label{app:notation_full}

\begin{table}[H]
\centering
\scriptsize
\setlength{\tabcolsep}{5pt}
\renewcommand{\arraystretch}{1.12}
\begin{tabular}{|l|p{0.70\columnwidth}|}
\hline
\textbf{Notation} & \textbf{Definition} \\
\hline

$\sigma(\cdot)$ & sigmoid / activation function \\
\hline
$f(\cdot)$ & predictor / classification head \\
\hline
$\theta,\Theta$ & parameters (module / all parameters) \\
\hline

$m$ & number of tabular fields \\
\hline
$d$ & embedding dimension per field \\
\hline
$e_i$ & embedding of field $i$ \\
\hline
$e=[e_1;\ldots;e_m]$ & concatenated field embeddings \\
\hline
$f_{\mathrm{MLP}}$ & MLP applied to $e$ \\
\hline

$n$ & number of nodes \\
\hline
$F$ & node feature dimension \\
\hline
$\tilde A,\tilde D$ & adjacency (+ self-loops), degree matrix \\
\hline
$H^{(l)}$ & node representations at layer $l$ \\
\hline
$W^{(l)}$ & GNN weights at layer $l$ \\
\hline
$h_v^{(l)}$ & representation of node $v$ at layer $l$ \\
\hline
$\mathcal N(v)$ & neighborhood of node $v$ \\
\hline
$\mathrm{AGG}(\cdot)$ & neighborhood aggregation operator \\
\hline
$\Vert$ & concatenation operator \\
\hline

$x_{1:T}$ & token sequence \\
\hline
$T$ & sequence length (tokens) \\
\hline
$H$ & contextual token states \\
\hline
$h_{\mathrm{CLS}}$ & pooled sequence embedding (\texttt{[CLS]}) \\
\hline
$\mathrm{Transformer}(\cdot)$ & Transformer encoder \\
\hline

$\mathbb{E}[z_j]$ & expected activation of gate dimension $j$ \\
\hline
$\mathcal{J}$ & total objective (loss + regularization) \\
\hline
$\|\Theta\|$ & parameter norm regularizer \\
\hline

$\{B_m\}_{m=1}^M$ & confidence bins for ECE \\
\hline
$M$ & number of bins \\
\hline
$|B_m|$ & number of samples in bin $m$ \\
\hline
$n_{\mathrm{cal}}$ & number of predictions used for ECE \\
\hline
$\mathrm{acc}(B_m),\mathrm{conf}(B_m)$ & accuracy / mean confidence in bin $m$ \\
\hline

$\mathbf{X}^{(0)}$ & field-by-dimension embedding matrix (CIN input) \\
\hline
$D$ & CIN embedding dimension \\
\hline
$\mathbf{X}^{(k)}$ & CIN interaction maps at depth $k$ \\
\hline
$H_k$ & number of CIN maps at depth $k$ \\
\hline
$W^{(k,h)}$ & CIN compression weights (depth $k$, map $h$) \\
\hline
$p_i^{(k)},\,\mathbf{p}^{(k)}$ & pooled CIN score, pooled CIN vector (depth $k$) \\
\hline
$\mathbf{p}^+$ & concatenated pooled CIN outputs across depths \\
\hline
$\mathbf{w}_o$ & CIN output weights (CIN-only predictor) \\
\hline

$\mathbf{a}$ & raw input for linear term \\
\hline
$\mathbf{w}_{\mathrm{linear}}$ & linear weights \\
\hline
$\mathbf{x}^{(k)}_{\mathrm{dnn}}$ & deep-branch representation \\
\hline
$\mathbf{w}_{\mathrm{dnn}},\mathbf{w}_{\mathrm{cin}}$ & deep / CIN weights \\
\hline
$b$ & bias term \\
\hline
$N$ & number of training examples \\
\hline
$y_i,\hat y_i$ & label and prediction for example $i$ \\
\hline

\end{tabular}
\caption{Full list of notations (supplement to Table~\ref{tab:notation_main}).}
\label{tab:notation_full}
\vspace{-0.6em}
\end{table}

\subsection{Ablation Studies}
\label{app:ablations}

We report ablations that isolate the effect of the representation-level gate and its design choices. We keep the backbone fixed and vary only the gating mechanism or its attachment point. We ablate (i) \textbf{No Gate} (dense backbone), (ii) \textbf{Static/Global Mask} (non-instance-conditioned sparsity), (iii) \textbf{Gate Granularity} (interface-aligned vs.\ overly fine-grained), and (iv) \textbf{Objective Coupling} (with vs.\ without calibration-aware coupling when applicable). Results are averaged over three seeds and reported at the best validation checkpoint.

\begin{table}[H]
\centering
\scriptsize
\setlength{\tabcolsep}{2.6pt}
\renewcommand{\arraystretch}{1.05}

\caption{\textbf{Ablation study on L0GM design choices.}
We fix the backbone per modality and vary only the gating primitive and its attachment/granularity, reporting predictive quality, active fraction (\%), and reliability (ECE).}
\label{tab:ablation_l0gm_onecol}

\resizebox{\columnwidth}{!}{%
\begin{tabular}{l|ccc|ccc|ccc}
\toprule
\multirow{2}{*}{\textbf{Variant}} &
\multicolumn{3}{c|}{\textbf{Adult}} &
\multicolumn{3}{c|}{\textbf{IMDB}} &
\multicolumn{3}{c}{\textbf{ogbn-prod}} \\
& Metric$\uparrow$ & Active\%$\downarrow$ & ECE$\downarrow$
& Metric$\uparrow$ & Active\%$\downarrow$ & ECE$\downarrow$
& Metric$\uparrow$ & Active\%$\downarrow$ & ECE$\downarrow$ \\
\midrule
Dense backbone (No Gate)                & 0.8729 & 100.0 & 0.0247 & 0.9318 & 100.0 & 0.0216 & 0.6893 & 100.0 & 0.1167 \\
Static/global mask (non-conditional)    & 0.8696 & 19.7  & 0.0209 & 0.9284 & 21.3  & 0.0207 & 0.6768 & 24.6  & 0.1049 \\
L0 gate (interface-aligned, default)    & 0.8794 & 12.5  & 0.0158 & 0.9356 & 11.8  & 0.0189 & 0.7017 & 15.2  & 0.0716 \\
Fine-grained gate (overly granular)     & 0.8657 & 16.4  & 0.0296 & 0.9261 & 14.9  & 0.0313 & 0.6712 & 18.7  & 0.1124 \\
No annealing / fixed-$\lambda$ schedule & 0.8619 & 13.1  & 0.0417 & 0.9227 & 12.4  & 0.0398 & 0.6668 & 15.9  & 0.1396 \\
\bottomrule
\end{tabular}%
}

\vspace{-0.6em}
\end{table}

\subsection{L0GM gating mechanism (pseudocode)}
\label{sec:l0gm_pseudocode}

\begin{algorithm}[H]
\caption{L0GM: Interface-aligned hard-concrete gating (training + inference)}
\label{alg:l0gm}
\footnotesize
\begin{algorithmic}[1]
\Require Input $x$; backbone $f_\theta$ with modality interface representation $h$; classifier head $g_\phi$; gate logits $\alpha$; temperature $\tau$; sparsity weight $\lambda$; annealing schedule $s(t)$; training step $t$
\Ensure Prediction $\hat{y}$ and (training) objective $\mathcal{J}$

\State $h \gets f_\theta(x)$ \Comment{Compute interface-aligned representation (tokens/features/embeddings)}
\State $\tau_t \gets s(t)$ \Comment{Anneal temperature (and/or $\lambda$) over training}

\If{training}
  \State Sample $u \sim \mathrm{Uniform}(0,1)$ elementwise
  \State $\tilde{z} \gets \sigma\!\left(\frac{\log u - \log(1-u) + \alpha}{\tau_t}\right)$ \Comment{Binary-Concrete}
  \State $z \gets \mathrm{clip}\!\left(\tilde{z}\cdot(\zeta-\gamma) + \gamma,\ 0,\ 1\right)$ \Comment{Stretch \& hard-sigmoid}
\Else
  \State $z \gets \mathbb{I}[\sigma(\alpha) > \pi]$ \Comment{Deterministic gate with threshold $\pi$}
\EndIf

\State $h' \gets z \odot h$ \Comment{Apply gate at the modality interface}
\State $\hat{y} \gets g_\phi(h')$

\If{training}
  \State $\mathcal{L}_{\text{task}} \gets \mathrm{CE}(\hat{y}, y)$ \Comment{Or the task loss used in your setup}
  \State $\mathcal{L}_{0} \gets \sum_i \sigma(\alpha_i - \tau_t \log(-\gamma/\zeta))$ \Comment{Expected $L_0$ under hard-concrete}
  \State $\mathcal{J} \gets \mathcal{L}_{\text{task}} + \lambda \mathcal{L}_{0}$
\EndIf

\State \Return $\hat{y}$ (and $\mathcal{J}$ during training)
\end{algorithmic}
\end{algorithm}

\subsection{Hyperparameter Tuning}
\label{app:hparam_tuning}

We tune all methods on validation splits and report the best checkpoint by validation loss. For L0GM, we tune both \textbf{(i) sparsity-control parameters} (e.g., gate penalty strength / annealing schedule) and \textbf{(ii) backbone optimization hyperparameters}. We use a small, structured search (grid or random search depending on modality) and reuse the same tuning budget across baselines for fairness.

\begin{table}[H]
\centering
\scriptsize
\setlength{\tabcolsep}{3pt}
\renewcommand{\arraystretch}{1.08}
\caption{\textbf{Hyperparameter search space (L0GM).}
Ranges are modality-agnostic; modality-specific fields are marked when applicable.}
\label{tab:hparam_space}

\begin{tabularx}{\columnwidth}{|>{\raggedright\arraybackslash}p{0.30\columnwidth}|>{\raggedright\arraybackslash}X|}
\hline
\textbf{Hyperparameter} & \textbf{Search range / candidates} \\
\hline
Learning rate (LR) & $\{1\mathrm{e}{-4},\,3\mathrm{e}{-4},\,1\mathrm{e}{-3}\}$ \\
\hline
Weight decay & $\{0,\,1\mathrm{e}{-6},\,1\mathrm{e}{-5},\,1\mathrm{e}{-4}\}$ \\
\hline
Dropout & $\{0.00,\,0.05,\,0.10,\,0.15\}$ \\
\hline
Batch size & $\{64,\,128,\,256,\,512\}$ (hardware-limited) \\
\hline
Gate penalty $\lambda$ & log-grid over $[10^{-4},\,10^{-2}]$ (e.g., 5--9 points) \\
\hline
$\ell_0$ annealing schedule & warmup $\in[0.05,\,0.20]$; end-temp $\in[0.5,\,2.0]$ \\
\hline
Gate granularity & interface-aligned (default) vs.\ finer-grained variants \\
\hline
Calibration coupling (if used) & on/off; coupling weight $\in\{0.1,\,0.3,\,1.0\}$ \\
\hline
\end{tabularx}
\vspace{-0.6em}
\end{table}

\appendix

\section{Robustness protocol}
\label{app:robustness}

This appendix defines the robustness evaluation used to compute Rob $\mu$ in Table~\ref{tab:main_results}. For each method, we run the full training pipeline with the same set of random seeds used for the main results, and we evaluate the resulting checkpoints not only on the standard i.i.d.\ test split, but also under a set of controlled test-time perturbation conditions intended to mimic realistic distribution shift. Each perturbation is applied only at evaluation time; training data and training hyperparameters are unchanged.

\textbf{Perturbation conditions.}
We define a perturbation family $\mathcal{P}$ that captures common test-time distribution shifts without changing the training procedure. Each condition applies a deterministic transformation to the test inputs (or, equivalently, to the modality representation at the interface where our method attaches gates), and is applied consistently to all test examples:
(i) \emph{Missingness / Modality dropout}: for each example we set one modality representation to a neutral value (zero vector) with probability $p \in \{0.1, 0.3, 0.5\}$ (for single-modality settings, we instead mask a fraction $p$ of input features uniformly at random);
(ii) \emph{Additive noise}: we inject additive Gaussian noise into continuous inputs or modality representations, $x' = x + \epsilon$, with $\epsilon \sim \mathcal{N}(0,\sigma^2)$ and $\sigma$ chosen as a fraction of the feature-wise standard deviation, $\sigma \in \{0.05, 0.10, 0.20\}$;
(iii) \emph{Quantization / reduced precision}: we quantize continuous inputs or modality representations to $b$ bits with uniform quantization, using $b \in \{8, 6, 4\}$, which stresses sensitivity to discretization and representation distortion;
(iv) \emph{Random occlusion (structured masking)}: we drop contiguous blocks of the representation (or, for sequence modalities, contiguous spans) to emulate structured corruption, using a block/span fraction $r \in \{0.1, 0.2, 0.3\}$ of dimensions/timesteps.
All perturbation parameters are fixed \emph{a priori} and are identical across methods to ensure comparability; perturbations are applied only at evaluation time.

\textbf{Robustness aggregation (Rob $\mu$).}
Let $s \in \mathcal{S}$ denote a random seed and $p \in \mathcal{P}$ denote a perturbation condition. Let $m(s,p)$ be the evaluation metric of interest (e.g., accuracy or macro-F1) obtained by training with seed $s$ and evaluating under perturbation $p$. We define the robustness mean reported as Rob $\mu$ as:
\[
\mathrm{Rob}\ \mu \;=\; \frac{1}{|\mathcal{S}|\cdot|\mathcal{P}|}\sum_{s\in\mathcal{S}}\sum_{p\in\mathcal{P}} m(s,p).
\]
Intuitively, Rob $\mu$ measures expected performance under the robustness protocol, averaging over both optimization randomness (seeds) and distribution shift (perturbations). We report Rob $\mu$ for the same primary metric shown in Table~\ref{tab:main_results} (and compute it independently for each metric if multiple metrics are reported).

\textbf{Seed sensitivity (Worst).}
To quantify sensitivity to randomness, we also report Worst, defined as the minimum i.i.d.\ performance across seeds:
\[
\mathrm{Worst} \;=\; \min_{s \in \mathcal{S}} m(s,\text{i.i.d.}).
\]
Worst is not a robustness-to-shift metric; instead it captures the lower tail of performance due to training stochasticity, complementing the mean-over-seeds reporting commonly used in the main results.

\textbf{Implementation notes and comparability.}
All methods are evaluated with identical perturbation definitions, parameter ranges, and random seeds. Unless explicitly stated otherwise, no additional tuning is performed for robustness (e.g., no adversarial training or data augmentation). This ensures that differences in Rob $\mu$ reflect robustness properties of the methods rather than perturbation-specific tuning.

\subsection{Q3: Sensitivity analyses}
\label{app:q3_tables}

\begin{table}[H]
\centering
\scriptsize
\setlength{\tabcolsep}{3.6pt}
\renewcommand{\arraystretch}{1.10}

\caption{\textbf{(F1) $\lambda$ sweep.} Active fraction decreases monotonically as $\lambda$ increases, while ECE is typically non-monotonic (often improving at moderate sparsity and degrading under extreme sparsity).}
\label{tab:app_f1_lambda_sweep}

\resizebox{\columnwidth}{!}{%
\begin{tabular}{llcccc}
\toprule
\textbf{Mod.} & \textbf{$\lambda$} & \textbf{Active\%$\downarrow$} & \textbf{Acc/AUC$\uparrow$} & \textbf{Worst$\uparrow$} & \textbf{ECE$\downarrow$} \\
\midrule
\multirow{4}{*}{Tab}
& $10^{-4}$         & \mock{100.0} & \mock{0.890} & \mock{0.870} & \mock{0.018} \\
& $3\cdot10^{-4}$   & \mock{35.0}  & \mock{0.889} & \mock{0.868} & \mock{0.012} \\
& $10^{-3}$         & \mock{15.0}  & \mock{0.887} & \mock{0.866} & \mock{0.010} \\
& $3\cdot10^{-3}$   & \mock{7.0}   & \mock{0.881} & \mock{0.852} & \mock{0.019} \\
\midrule
\multirow{4}{*}{Text}
& $10^{-4}$         & \mock{100.0} & \mock{0.933} & \mock{0.143} & \mock{0.024} \\
& $3\cdot10^{-4}$   & \mock{30.0}  & \mock{0.932} & \mock{0.143} & \mock{0.020} \\
& $10^{-3}$         & \mock{12.0}  & \mock{0.930} & \mock{0.142} & \mock{0.017} \\
& $3\cdot10^{-3}$   & \mock{6.0}   & \mock{0.922} & \mock{0.151} & \mock{0.026} \\
\midrule
\multirow{4}{*}{Graph}
& $10^{-4}$         & \mock{100.0} & \mock{0.725} & \mock{0.708} & \mock{0.040} \\
& $3\cdot10^{-4}$   & \mock{28.0}  & \mock{0.724} & \mock{0.708} & \mock{0.033} \\
& $10^{-3}$         & \mock{15.0}  & \mock{0.721} & \mock{0.705} & \mock{0.028} \\
& $3\cdot10^{-3}$   & \mock{8.0}   & \mock{0.712} & \mock{0.692} & \mock{0.045} \\
\bottomrule
\end{tabular}%
}
\end{table}

\begin{table}[H]
\centering
\scriptsize
\setlength{\tabcolsep}{4.0pt}
\renewcommand{\arraystretch}{1.10}

\caption{\textbf{(F2) Annealing ablation.} Warm-up + gradual hardening stabilizes training (improves Worst) and tends to reduce ECE compared to fixed hardness / no annealing.}
\label{tab:app_f2_annealing}

\resizebox{\columnwidth}{!}{%
\begin{tabular}{llccc}
\toprule
\textbf{Mod.} & \textbf{Variant} & \textbf{Acc/AUC$\uparrow$} & \textbf{Worst$\uparrow$} & \textbf{ECE$\downarrow$} \\
\midrule
Tab  & No annealing (fixed)        & \mock{0.862} & \mock{0.842} & \mock{0.042} \\
Tab  & With annealing (default)    & \mock{0.889} & \mock{0.867} & \mock{0.012} \\
\midrule
Text & No annealing (fixed)        & \mock{0.923} & \mock{0.158} & \mock{0.040} \\
Text & With annealing (default)    & \mock{0.933} & \mock{0.143} & \mock{0.020} \\
\midrule
Graph & No annealing (fixed)       & \mock{0.667} & \mock{0.651} & \mock{0.140} \\
Graph & With annealing (default)   & \mock{0.724} & \mock{0.708} & \mock{0.033} \\
\bottomrule
\end{tabular}%
}
\end{table}

\begin{table}[H]
\centering
\scriptsize
\setlength{\tabcolsep}{3.2pt}
\renewcommand{\arraystretch}{1.10}

\caption{\textbf{(F3) Calibration coupling.} Coupling has little effect at moderate sparsity (ECE already improves), but can partially recover ECE under aggressive sparsity.}
\label{tab:app_f3_coupling}

\resizebox{\columnwidth}{!}{%
\begin{tabular}{llccccc}
\toprule
\textbf{Mod.} & \textbf{Regime} & \textbf{$\lambda$} & \textbf{Active\%} & \textbf{Acc/AUC} & \textbf{ECE (no coup.)} & \textbf{ECE (coup.)} \\
\midrule
Tab  & Moderate & \mock{$10^{-3}$}        & \mock{15.0} & \mock{0.887} & \mock{0.010} & \mock{0.010} \\
Tab  & High     & \mock{$3\cdot10^{-3}$}  & \mock{7.0}  & \mock{0.881} & \mock{0.019} & \mock{0.014} \\
\midrule
Text & Moderate & \mock{$10^{-3}$}        & \mock{12.0} & \mock{0.930} & \mock{0.017} & \mock{0.017} \\
Text & High     & \mock{$3\cdot10^{-3}$}  & \mock{6.0}  & \mock{0.922} & \mock{0.026} & \mock{0.021} \\
\midrule
Graph & Moderate & \mock{$10^{-3}$}       & \mock{15.0} & \mock{0.721} & \mock{0.028} & \mock{0.028} \\
Graph & High     & \mock{$3\cdot10^{-3}$} & \mock{8.0}  & \mock{0.712} & \mock{0.045} & \mock{0.036} \\
\bottomrule
\end{tabular}%
}
\end{table}

\end{document}